\documentclass[letterpaper,twocolumn,10pt]{article}
\usepackage{usenix-2020-09}
\usepackage{tikz}
\usepackage{amsmath}
\usepackage{amsfonts}
\usepackage{amssymb}
\usepackage{amsthm}
\usepackage{authblk}
\usepackage{booktabs}
\usepackage{mathtools}
\usepackage{xspace}
\usepackage{listings}
\usepackage{caption}
\usepackage{subcaption}
\usepackage{tablefootnote}
\usepackage{multicol}
\usepackage{multirow}
\usepackage{pifont}
\usepackage{siunitx}
\usepackage[nameinlink]{cleveref}
\usepackage[linesnumbered,ruled,vlined]{algorithm2e}

\makeatletter
\renewcommand{\paragraph}{%
  \@startsection{paragraph}{4}%
  {\z@}{0.5ex \@plus 1ex \@minus .2ex}{-1em}%
  {\normalfont\normalsize\bfseries}%
}
\makeatother

\AtBeginDocument{%
  }

\newcommand{\systemname}{Lamina\xspace}

\newcommand{\techname}{model-attention disaggregation\xspace}
\DeclareMathOperator{\softmax}{softmax} 
\DeclareMathOperator{\mtime}{MTIME} 
\DeclareMathOperator{\atime}{ATIME} 
\newcommand{\DOP}{\text{DOP}}

\begin{document}

\title{Efficient Heterogeneous Large Language Model Decoding \\ with Model-Attention Disaggregation}
\date{}
\author[1]{Shaoyuan Chen}
\author[2]{Wencong Xiao}
\author[1]{Yutong Lin}
\author[1]{Mingxing Zhang}
\author[1]{Yingdi Shan}
\author[1]{Jinlei Jiang}
\author[1]{Kang Chen}
\author[1]{Yongwei Wu}

\affil[1]{Tsinghua University}
\affil[2]{ByteDance}
% \author{Shaoyuan Chen\textsuperscript{1} \and Wencong Xiao\textsuperscript{2} \and Yutong Lin\textsuperscript{1} \and Mingxing Zhang\textsuperscript{1} \and Yingdi Shan\textsuperscript{1} \and Jinlei Jiang\textsuperscript{1} \and Kang Chen\textsuperscript{1} \and Yongwei Wu\textsuperscript{1}}

\maketitle

% \begin{center}
%     \textsuperscript{1} Tsinghua University \quad \textsuperscript{2} ByteDance
% \end{center}

\begin{abstract}
Transformer-based large language models (LLMs) exhibit impressive performance in generative tasks but also introduce significant challenges in real-world serving due to inefficient use of the expensive, computation-optimized accelerators. Although disaggregated serving architectures have been proposed to split different phases of LLM inference, the efficiency of decoding phase is still low. This is caused by the varying resource demands of different operators in the transformer-based LLMs. Specifically, the attention operator is memory-intensive, exhibiting a memory access pattern that clashes with the strengths of modern accelerators, especially for long context requests.

To enhance the efficiency of LLM decoding, we introduce \techname. This approach leverages a collection of cheap, memory-optimized devices for the attention operator while still utilizing high-end accelerators for other parts of the model. This heterogeneous setup ensures that each component is tailored to its specific workload, maximizing overall performance and cost efficiency. Our comprehensive analysis and experiments confirm the viability of splitting the attention computation over multiple devices. Also, the communication bandwidth required between heterogeneous devices proves to be manageable with prevalent networking technologies. To further validate our theory, we develop and deploy \systemname, an LLM inference system that incorporates \techname in a distributed heterogeneous cluster. Experimental results indicate that \systemname can provide $16.1 \sim 90.1\%$ higher estimated throughput than existing solutions with similar costs.
\end{abstract}

% \keywords{Do, Not, Us, This, Code, Put, the, Correct, Terms, for, Your, Paper}

\pagestyle{plain}
    
\section{Introduction} \label{sec:introduction}
\subsection{Motivation}
Disaggregated serving architectures for large language models (LLMs) \cite{zhong2024distserve,patel2024splitwise,qin2024mooncake} have recently emerged as efficient frameworks for handling generative inference requests. The core concept of disaggregation involves allocating separate resources for different tasks to improve resource utilization. This approach aligns perfectly with LLM processing, which can be divided into two distinct phases. The first phase, known as the prefill phase, processes all input tokens from the prompt in parallel and is computation-bound. The second phase, i.e., the decode phase, generates the output tokens one after another, and is typically memory-bound.

Splitting the two phases of inference reduces interference between different requests and allows for more flexible parallel configurations for the two phases. To better leverage the differing characteristics of each phase, several methods propose using heterogeneous hardware to reduce the cost of disaggregated serving \cite{zhong2024distserve,anonymous2024hexgen}. Specifically, flagship all-rounder GPUs like NVIDIA H100 integrate high-performance computational units and high-bandwidth memory (HBM) within a single package, delivering good performance for LLM inference. However, as shown in \autoref{tab:accelerator-specs}, specialized accelerators optimized for either computation or bandwidth can be significantly cheaper than the H100 in terms of TFLOPS per dollar/watt (e.g., TPU v6e) or bandwidth per dollar/watt (e.g., NVIDIA H20), but not both. This cost disparity arises because all-rounder GPUs combine powerful computation units, HBM controllers, and high-bandwidth internal buses within a single chip. Such integration leads to larger die sizes and increased transistor counts, posing additional challenges for chip designing, packaging, and thermal management \cite{gholami2024ai,huang2011scaling,yang2024challenges}, all of which drive up the design and manufacturing cost. % Also, as new GPU generations are released approximately every 24 months, individual GPU types remain in use for much longer periods. As a result, heterogeneous accelerators are standard in data centers. Based on this situation, many prior works have proposed using different accelerators for different phases, selecting the best TFLOPS per dollar for the prefill phase and the best bandwidth per dollar for the decoding phase. 

\begin{table}[htbp]
    \begin{center}
    \caption{H100, H20, and TPU v6e specifications.} \label{tab:accelerator-specs}
    \small
    \begin{tabular}{c|ccc}
    \toprule
         & \textbf{H100} &  \textbf{H20} & \textbf{TPU v6e} \cite{gcp-tpu-v6e}  \\
    \midrule
        \textbf{BF16 TFLOPs} & 989 & 148 & 918 \\
        \textbf{Memory capacity} & 80 GB & 96 GB & 32 GB\\ 
        \textbf{Memory bandwidth} & 3.35 TB/s & 4.0 TB/s & 1.64 TB/s \\
        \textbf{Power rating} & 700 W & 400 W & unlisted \\
        % 下面两行后面会引用到
        \textbf{Inter-chip bandwidth} & 450 GB/s & 450 GB/s & 448 GB/s \\ 
        \textbf{Network bandwidth} & 400 Gbps & 400 Gbps & 200 Gbps \\
        \textbf{Price per chip \cite{googleCloudComputing}} & \$11.06/hr 
        & \$4.63/hr * & \$2.70/hr  \\
    \bottomrule
    \end{tabular}
    \end{center}
    \footnotesize{*: As H20 is not readily available on cloud service providers, the listed price is estimated using the relative complete system cost against H100.}
\end{table}

According to our analyses and experiments, while the separation of resources works well for the prefill nodes, we identified significant inefficiencies in the decoding phase. For instance, as analyzed in \autoref{sec:background}, the computation resource utilization is often below 20\% when serving the LLaMA3-70B model with H100. This is primarily due to the limited GPU memory size, which cannot accommodate the large aggregated KV cache for large batches, as well as the low arithmetic intensity of the attention operators.

A detailed examination reveals that the decoding phase mainly comprises two types of operators, each facing distinct resource bottlenecks. Linear transformations, including QKVO projections and feedforward networks, are implemented with \textit{generalized matrix-matrix multiplications} (GEMMs). Since all requests multiply with the same parameter matrices in these operators, processing multiple requests in batch can avoid repeated parameter loads from memory, making these operators primarily computation-bound. In contrast, the self-attention operator is memory-bound. This pivotal operator requires each request to read its own, distinct KV cache, resulting in a  \textit{batched generalized matrix-vector multiplication} (BGEMV) pattern. Increasing batch sizes does not improve the computation resource utilization but places additional pressure on the already limited memory capacity.

\subsection{Our Contributions}

\begin{figure}[ht]
    \centering
    \includegraphics[width=\linewidth]{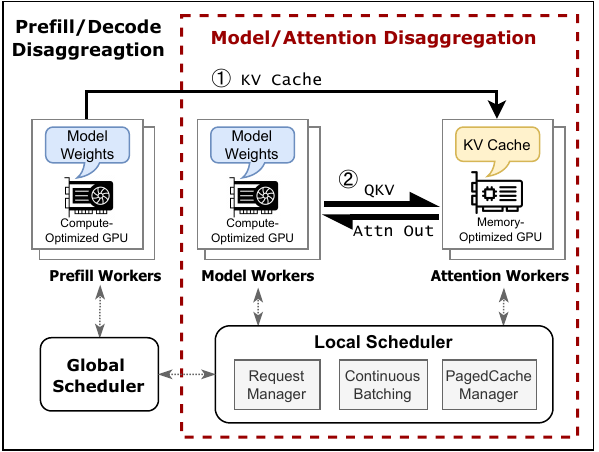}
    \caption{The disaggregated architecture of LLM serving.}
    \label{fig:ao-overview}
\end{figure}

In light of the above findings, we propose an innovative concept called \textbf{\techname}, as illustrated in Figure \ref{fig:ao-overview}. This approach involves further disaggregating the decoding phase by creating two pools of heterogeneous accelerators: one optimized for computational power and the other for memory resources. We use the memory-optimized accelerators to store the KV caches and process self-attention operators, while the computation-optimized devices handle all other operators. By choosing the most suitable devices for each kind of operators, this architecture further increases hardware utilization and leads to better overall performance. Moreover, different LLMs and workloads present varying computation and memory resource requirements. Homogeneous accelerator solutions, however, can only provide \textbf{a fixed ratio of computation and memory resources}, which can result in resource wastage. For instance, as context lengths increase, the memory capacity needed to store the KV cache expands accordingly; with a fixed resource ratio, a substantial portion of computational resources remains underutilized when processing requests with long contexts. By pooling heterogeneous accelerators, we can adjust the number of each kind of accelerators to better match the LLM and workload and hence improve resource utilization. 

The primary challenge associated with attention offloading arises from the substantial communication demands between heterogeneous accelerators when sending and receiving the inputs and outputs of self-attention operators. Unlike the original prefill-decode disaggregation, where the KV cache is transferred only once between the prefill nodes and the decode nodes, our model-attention disaggregation architecture requires inter-GPU communication for every layer of the model. Even worse, communication between heterogeneous GPUs must rely on \textit{data center networks} (DCNs), such as Ethernet and InfiniBand, which provide only \char`\~10\% of the bandwidth of \textit{inter-chip interconnects} (ICIs) like NVLink between homogeneous GPUs. If not handled properly, this frequent communication would introduce high network round-trip times (RTTs) to the token generation latency, worsening the user experience. 

To assess the practicality of our novel disaggregated architecture, we first conduct a detailed quantitative study indicating that these concerns are manageable in the context of LLM inference. In \autoref{subsec:distributed-ao}, we provide profiling and analysis to determine the minimum bandwidth threshold between different accelerator pools. Our findings reveal that 200/400Gbps DCNs, widely deployed in current AI-oriented data centers, suffice for attention offloading. However, this can only be achieved if the inter-GPU communication is carefully implemented and optimized, which is not possible for off-the-shelf communication libraries such as NCCL or Gloo.

To realize the idea of model/attention disaggregation, we implement two specific techniques to reduce the networking overhead. First, we designed and deployed a fully host-bypassed network stack. Leveraging PCIe P2P capabilities, this revamped network stack enables GPUs to directly talk with network interface cards (NICs), eliminating the need for host CPU synchronization and involvement for network transmissions. The network data is also directly read from and written to GPU memory without passing through host memory. Additionally, we developed an automated model converter. This converter splits the model computation graph into slices, interleaved with attention operators. It also reorders the operators and coordinates the computation and communication pipelines, enabling effective overlapping of communication and computation tasks. 

Moreover, with model-attention disaggregation, running the inference process with only a single batch results in underutilization of resources, as the memory device remains idle when the computation device is active, and vice versa. To address this inefficiency and resource wastage, we introduce staggered pipelining, an advanced technique that increases the hardware utilization. With staggered pipelining, we run multiple batches concurrently and optimize the workflow to ensure that both the computation and memory devices are working simultaneously, minimizing resource waste and maximizing system performance.

To validate our analysis, we develop and evaluate \systemname, a distributed heterogeneous LLM inference system with \techname. We also conduct extensive evaluations to mirror the real-world LLM services with a heterogeneous cluster made up of H100 and H20 GPUs, tested with various models and request traces collected from the production environments of LLM service providers. Experimental results that our system can achieve up to $16.1\sim90.1\%$ higher throughput with similar hardware cost than existing solutions. Although \systemname experiences a slightly larger latency than homogeneous solutions for the larger ($2.39\times$ on average) batch sizes and additional networking and scheduling costs,  the latency is still within the SLO of online interactive LLM services.

\begin{figure*}[t]
    \centering
    \vspace{1mm}
    \includegraphics[width=0.96\linewidth]{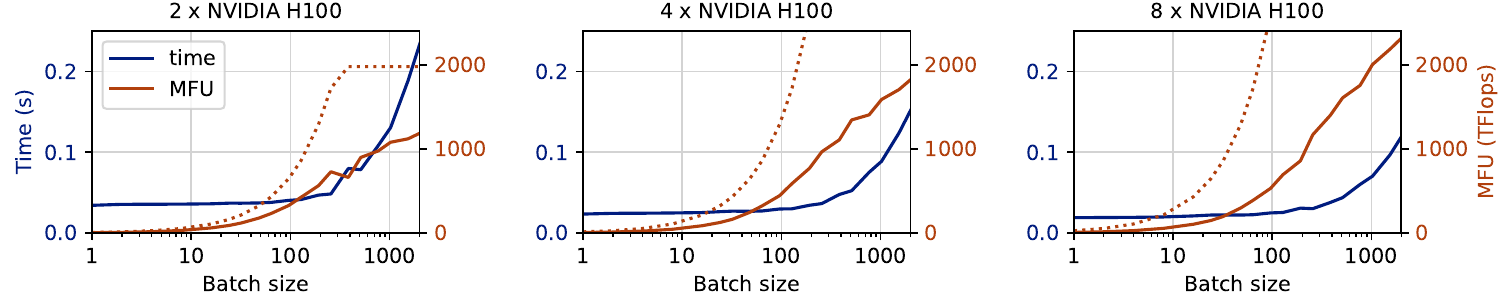}
    \caption{Measured time consumption and MFU of non-attention operators in LLaMA3-70B during one decode iteration. Results with different tensor parallelisms are presented. The dotted lines indicate the projected values using the roofline model.}
    \label{fig:measured-llama70b}
\end{figure*}

\section{Background: The Underutilization of GPUs in LLM Decoding} \label{sec:background}
% Generative AI services, like coding assistants and chatbots, rely on large language models (LLMs) to produce meaningful results. Given an input text (prompt) represented as a sequence of \emph{tokens} $(w_1, w_2, \dots, w_n)$, an LLM predicts the probability distribution of the next token $P(w_{n+1}|w_1, w_2, \dots, w_n)$. To generate a continuous stream of text, this process is repeated iteratively. At each step, the LLM predicts the probabilities for the subsequent token, samples one based on these probabilities, and appends it to the growing text sequence. This approach, known as \emph{autoregressive generation}, continues until an end-of-sequence (EOS) token is generated or the generated text reaches a predefined length limit. % Through this mechanism, LLMs e\textbf{}nable generative AI services to produce coherent and contextually relevant responses to user prompts.

% To generate a sequence of text, we may iteratively run the LLM, sample from the predicted probability distribution, and append the sampled token to the input text. This procedure, called \emph{autoregressive generation}, is repeated until an end-of-sequence (EOS) token is generated or the length of the text reaches a predefined threshold.

To comprehensively understand the challenges and limitations present in current LLM decoding implementation with homogeneous hardware, 
this section will provide a detailed performance analysis of LLM decoding with LLaMA3-70B model as a representative LLM.
The specific notations used in this analysis are explained in \autoref{tab:performance-analysis-parameters}.

\begin{table}[htbp]
    \centering
    \small
    \caption{Notations used in the performance analysis. The values for LLaMA3-70B are also presented.}
    \label{tab:performance-analysis-parameters}
    \begin{tabular}{ccc} 
        \toprule
        \textbf{Parameter} & \textbf{Description}  & \textbf{Typical Value} \\ \midrule
        $N$ & Number of parameters in LLM. & 70 billion \\
        $d$ & Hidden dimension. & 8192 \\ 
        $L$ & Layers of the LLM. & 80 \\
        $G$ & GQA group size. & 8 \\
        $e$ & Bytes per element. & 2 \\ \midrule
        $B$ & Batch size. & $ 1 \sim 1024 $ \\ 
        $l$ & Sequence length. & $128 \sim 32768$ \\ \bottomrule
    \end{tabular} 
\end{table}

\subsection{Preliminaries}
% Most contemporary LLMs are based on the \emph{transformers} \cite{attention-is-all-you-need} architecture. A transformer-based LLM first maps each input token to a word embedding of dimension $d$. The word embeddings then go through a sequence of transformer blocks. Finally, the outputs of the last transformer block are multiplied by a sampling matrix to obtain the predicted likelihoods of the next tokens. 

% \paragraph{Transformer-based LLMs.} 
% Most contemporary LLMs are based on the \emph{transformers} \cite{attention-is-all-you-need} architecture. A transformer-based LLM first maps each input token to a word embedding of dimension $d_\text{embd}$. The word embeddings then go through a sequence of transformer blocks. Finally, the outputs of the last transformer block are multiplied by a sampling matrix to obtain the predicted likelihoods of the next tokens. 

% \begin{figure}[htbp]
%     \centering
%     \includegraphics[width=0.36\textwidth]{figures/llm-block-computation-graph.pdf}
%     % 这边就用GPT的例子，LLaMA中有GatedSiLU比较复杂
%     \caption{Computation graph of one LLM block.}
%     \label{fig:llm-block}
% \end{figure}

Modern large language models (LLMs) primarily rely on the transformer architecture \cite{attention-is-all-you-need}. In a transformer-based LLM, each input token is first mapped to a word embedding of dimension $d$. These embeddings then pass through a series of transformer blocks. The final output embeddings are multiplied by a sampling matrix to generate the predicted likelihoods for the next token.

Within each transformer block, the input embeddings are projected into three distinct vectors: query ($q_i$), key ($k_i$), and value ($v_i$), all of which have the same dimension $d$ as hidden states. These vectors are processed through an \emph{attention operator} to compute attention scores. The attention scores are then weighted by a matrix $W_{\text{out}}$ to produce the output embeddings $y_i$ of the attention layer.

\[
\begin{aligned}
    &q_i = W_q x_i, \quad k_i = W_k x_i, \quad v_i = W_v x_i,  \\
    &a_i = \sum_{j=1}^n \softmax \left(\frac{q_i^\top k_j}{\sqrt{d}}\right) v_j, \quad \star \\
    &y_i = W_{\text{out}} a_i .
\end{aligned}
\]

The output $y_i$ is then passed through a feedforward network that scales it into an intermediate vector space, followed by another matrix multiplication to scale it back:

\[
    x'_i = W_{\text{proj}} \cdot f_{\text{act}}\left(W_{\text{fc}} \cdot y_i\right) .
\]

Although the transformer block involves various transformations, there are actually only two kind of computationally expensive operations, which are the attention operator (denoted by $\star$ in the equations) and the other matrix projection steps.
Thus, in the following of this section,  we will conduct a quantitative analysis based on the roofline model \cite{williams-roofline} and experimental measurements to evaluate these two kinds of operators. This analysis will highlight the differing characteristics of attention and non-attention operators during the decoding phase, which explains why current LLM decoding implementations with homogeneous hardware often lead to underutilization of GPUs, thus motivating the need for heterogeneous architectures.

\subsection{Hardware Underutilization}

\subsubsection{The Underutilization in Non-Attention Operators} 

To improve GPU utilization in LLM decoding, continuous batching is widely adopted \cite{cellular-batching, lazy-batching, e-batch}. By processing multiple inputs concurrently, the model parameters in GPU memory can be reused, making the workload more computation-intensive. For a batch of $B$ requests, the non-attention operator requires approximately $2NB$ floating-point operations. Additionally, these operators involve loading model parameters $eN$ and reading/writing a total of $2eBd$ input and output data from GPU memory. The resulting arithmetic intensity, $\frac{2NB}{e(N+2Bd)}$, increases rapidly with larger batch sizes.

\autoref{fig:measured-llama70b} shows the latency and memory throughput utilization (MFU) of non-attention operators in LLaMA3-70B, measured on an NVIDIA H100 GPU, alongside projections based on the roofline model. For small batch sizes (less than 100), the workload is bandwidth-bound, with latency predominantly caused by accessing model parameters from GPU memory. In this regime, the MFU remains below 20\%, indicating significant underutilization of computational resources. As the batch size increases, the workload transitions to being computation-bound, with an increase in latency. To optimize GPU resource utilization, larger batch sizes are preferred. But, achieving this is often constrained by the limited VRAM capacity, which cannot accommodate the required KV cache size, a limitation discussed in detail later.

\subsubsection{The Underutilization in Attention Operators} 

Different from the weight matrix projection operators, the attention operator, when processing a batch of requestss still performs a batched matrix-vector multiplication, where each query accesses and processes its own KV cache. As a result, the arithmetic intensity of the attention operator remains constant, irrespective of the batch size. This behavior makes attention operations memory-bound, and increasing the batch size does not improve resource utilization. More recent models have adopt \emph{grouped-query attention} (GQA), which splits $q_i$ into a group of $G$ independent queries and reduce the size of $k_i$ and $v_i$ by a factor of $G$. Each query goes through the attention computation with the same $k_i$ and $v_i$ and the outputs are simply concatenated. With GQA, the arithmetic intensity of attention operators is increased $G$ times, but is still quite low compared with other operators.

\begin{figure}[htbp]
    \centering
    \includegraphics[width=\linewidth]{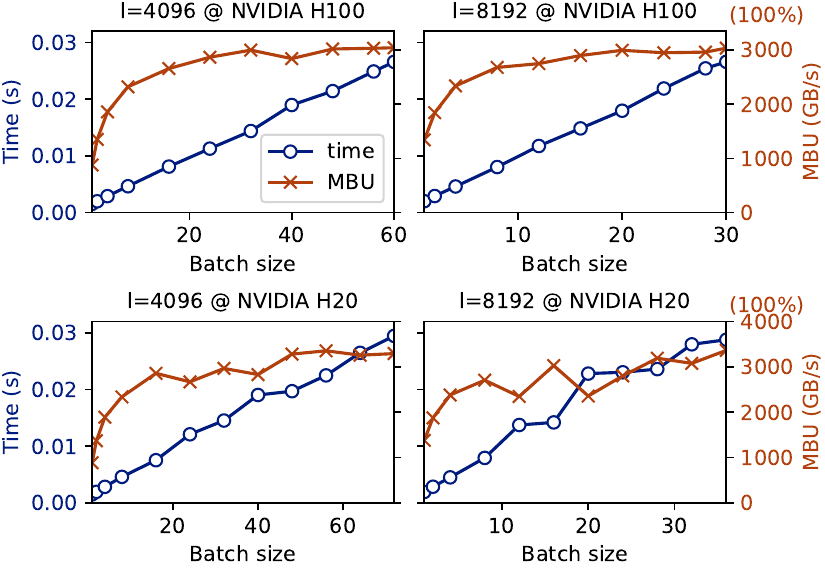}
    \caption{Measured time consumption and model bandwidth utilization (MBU) of attention operators in LLaMA3-70B during one decode iteration. Results with different sequence lengths and hardware configurations are presented.}
    \label{fig:attn-mbu}
\end{figure}

As shown in \autoref{fig:attn-mbu}, the bandwidth utilization of attention operators remains above 70\% even for small batch sizes, such as 20. This holds true even on memory-specialized accelerators like H20, which delivers only 15\% of the TFLOPs of the H100. However, the batch size achievable for attention operations is constrained by GPU memory capacity, particularly due to the high memory demand of KV caches for longer context lengths. For example, with a context length of 8192, the full memory of an H100 can only hold KV caches for about 30 requests, with the actual number being lower due to memory used by model weights. Consequently, the limited batch size for attention operations becomes a key bottleneck, preventing efficient utilization of computational resources for non-attention operations during the decoding phase.

\section{Model-Attention Disaggregation} \label{sec:offloading}

\subsection{Overview} \label{subsec:distributed-ao}

Current LLM serving systems often employ the same hardware for both attention and non-attention operators during the decode phase. However, our analysis reveals that this homogeneous approach leads to suboptimal resource utilization for both types of operators, due to the following reasons:
\begin{itemize}
\item \textbf{Attention operators} demonstrate low arithmetic intensity, as each value retrieved from the KV cache participates in only a limited number of computations. Given the disparity between memory bandwidth and computing power in modern high-performance accelerators, which favor high arithmetic intensity for efficient resource utilization, these operators tend to underutilize the computation resources of advanced GPUs.
\item For \textbf{non-attention operators}, while increasing the batch size could potentially enhance hardware utilization, this also results in a corresponding increase in the KV cache, which may exceed the available memory capacity. Consequently, to prevent memory overflow, the batch size is often kept small, which also leads to inefficient hardware utilization because of low arithmetic intensity.
\end{itemize}

To address the above limitations of homogeneous decoding solutions, we propose the \textbf{\techname} architecture, which uses memory-specialized accelerators to store KV caches and compute the attention operators; the non-attention operators are still executed on original accelerators. A \techname system can use multiple devices of each kind to provide different \textit{degrees of parallelism} (DOPs). If we use $a$ GPUs for non-attention operators and $b$ memory-optimized GPUs for attention operators, we denote the DOP as $(a, b)$.

By leveraging the cheaper memory-optimized devices, we can make larger batch sizes due to the extended memory capacities to store the KV caches, hence increasing the arithmetic intensity and promoting the hardware utilization of non-attention operators. Moreover, as the attention computation are moved to memory-optimized devices, we avoid wasting precious computation resources of high-end GPUs.

\begin{figure}
    \centering
    \begin{subfigure}[b]{0.48\linewidth}
        \includegraphics[width=\textwidth]{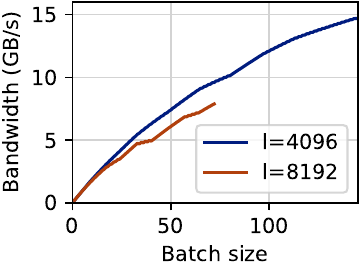}
        \caption{$\DOP=(2,2)$}
    \end{subfigure}
    \hfill
    \begin{subfigure}[b]{0.48\linewidth}
        \includegraphics[width=\textwidth]{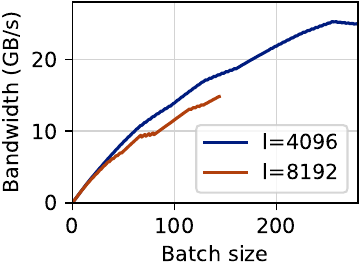}
        \caption{$\DOP=(2,4)$}
    \end{subfigure}
    \caption{The required network bandwidth for decoding LLaMA3-70B using attention offloading with H100 and H20, with at most $20\%$ latency slow-down for network overhead. }
    \label{fig:required-bandwidth}
\end{figure}

One potential obstacle in implementing attention offloading lies in the necessity of data transmission between heterogeneous accelerators for each layer of the model, which could encounter the communication wall problem and increase the end-to-end decoding latency. We conduct a quantitative analysis to determine the required interconnect bandwidth for such transfers. Say we run one iteration with batch size $B$, and we can afford $\alpha \times$ more latency for the networking overhead, the minimum interconnect bandwidth required can thus be calculated as 
\begin{align*}
    \text{minimum bandwidth} =& \frac{\text{size of data to transmit}}{\alpha \cdot \text{computation time}} \\
    =& \frac{(2+2/G)edBL}{\alpha [\mtime(B) + \atime(B, l)]} 
\end{align*}
where $\mtime(B)$ and $\atime(B,l)$ is running time of non-attention and attention operators at batch size $B$ and sequence length $l$, respectively, and they can be measured experimentally. The estimated minimum bandwidths required for different batch sizes, when $\alpha=0.2$, are calculated and presented in \autoref{fig:required-bandwidth}.

As evident from the data presented, the required interconnect bandwidth does not exceed 30 GB/s, even when dealing with batch sizes as high as 300. This bandwidth demand can be easily met by networking technologies like 400Gbps Ethernet. Indeed, contemporary data centers already fulfill this requirement, where each GPU is typically equipped with an exclusive 400Gbps NIC to provide sufficient networking bandwidth for LLM training.

For memory devices, the identical interconnection bandwidth is also necessary to communicate with computational devices. Since we employ a collection of more economical yet less powerful memory devices to collaboratively compute attention, the communication bandwidth needed for each individual device is significantly smaller. Consequently, we can choose to either equip each device with a less powerful NIC or install a single shared 400Gbps NIC to serve multiple memory devices.

\subsection{Practical Challenges}

While \techname promises potential benefits in improving LLM decoding efficiency, it also introduces a set of formidable practical challenges. We discuss some of these challenges below.

\paragraph{Frequent network communications.} By separating the attention operator from computation-optimized devices to memory-optimized devices, we introduce cross-machine data communications within each model layer. Even though the interconnect bandwidth in existing data centers is sufficient for attention offloading, we found that networking latency might still be a problem for efficient LLM decoding. With attention offloading, we have \textbf{layer-wise data transfer} between GPUs on different nodes, which may be up to thousands round-trips per second. These frequent network transfers might significantly increase the decoding time due to the accumulated network latencies. Hence, we need a refurnished, latency-optimized, GPU-aware networking stack for optimal performance of \techname.

\paragraph{Software engineering challenges.} With \techname, we are moving the execution of \textit{attention operator}, an intermediate operation of the transformer block, to other devices. This requires complicated and destructive modifications to the existing LLM codebase. Specifically, we have to dissect the models into separate slices that do not align with the modular structure of the transformer-based LLMs. This process is not only labor-intensive and error-prone but also significantly increases maintenance complexity. Hence, automated tools to help slice the models and perform relevant optimizations are highly desirable.

\paragraph{Difficult execution overlapping.} In a heterogeneous disaggregated system, various devices such as compute-optimized GPUs, memory-optimized GPUs, and NICs can be utilized simultaneously. Hence, we might achieve significant execution time reduction if the execution of operations occupying different devices could be overlapped. However, in the transformers architectures of current LLMs, attention operators and model operators are tightly interleaved in a sequential manner, with the output of one operator being transmitted over the network for the input of the other. Consequently, operations that depend on distinct hardware resources cannot be effectively overlapped in time, leading to considerable resource underutilization. Therefore, careful orchestration of operations on various devices and efficient design of task pipelines are required to promote execution overlapping and increase resource utilization.

\section{System Design} \label{sec:lamina}
% \begin{figure*}[htbp]
%     \centering
%      \begin{subfigure}[b]{0.39\textwidth}
%          \centering
%          \includegraphics[width=\textwidth]{figures/system-overview.pdf}
%          \caption{\systemname system overview.}
%          \label{fig:lamina-architecture:overview}
%      \end{subfigure}
%      \hfill
%      \begin{subfigure}[b]{0.59\textwidth}
%          \centering
%          \includegraphics[width=\textwidth]{figures/pipeline-stage-architecture.pdf}
%          \caption{Illustration of one pipeline stage.}
%          \label{fig:lamina-architecture:pipeline}
%      \end{subfigure}
%     \caption{System architecture of \systemname.}
%     \label{fig:lamina-architecture}
% \end{figure*}

We build \systemname, a distributed heterogeneous LLM decoding system that implements \techname and solves the related challenges. \systemname employs two kinds of acceleration devices: memory devices are used for storing KV cache and computing the attention operator, and computation devices are used for storing model parameters and computing other parts of the model. These two kinds of devices are interconnected with high-speed DCN, e.g., Infiniband or Ethernet. % These devices can either be co-located within the same physical machine or distributed across a cluster of nodes.

\subsection{Fully Host-Bypassed Network Stack}

% \begin{figure*}[htbp]
%     \centering
%     \includegraphics[width=0.7\linewidth]{figures/lamina-network.pdf}
%     \caption{Networking workflow} \label{fig:distributed-gpu-networking}
% \end{figure*}

The communication between GPUs across different nodes, often utilizing RDMA technologies, is a complex process that requires the coordination of multiple system agents, including the CPU, GPU, and NIC. To reduce GPU-aware networking overhead, GPUDirect RDMA (GDR) \cite{nvidiaGPUDirectRDMA} is developed to allow the RDMA-capable NIC (RNIC) to directly access GPU memory. This eliminates the need for host memory as an intermediate buffer, thereby enhancing both network latency and bandwidth. However, the control path still requires CPU involvement and includes several steps, all of which lie on the critical path and contribute to network latency. Specifically, when transferring data using GPUDirect RDMA, the following steps are performed:
\begin{enumerate} \itemsep0em 
    \item The local CPU waits for all prior GPU kernels to complete, ensuring the data to be transmitted is ready. \label{enum:rdma:first}
    \item The local CPU submits a send \textit{work request} (WR) to the RNIC. 
    \item The local RNIC processes the send WR, fetching the data from GPU memory and transmitting it over the physical network link.
    \item The remote RNIC receives the data and writes it to the GPU memory.
    \item The remote CPU waits for the RDMA receive operation to complete. \label{enum:rdma:lastbutone}
    \item The remote CPU launches the subsequent GPU kernels.
\end{enumerate}
Based on our experimental results, steps \ref{enum:rdma:first} through \ref{enum:rdma:lastbutone} may incur a latency of \qtyrange[range-units=single,range-phrase=--]{60}{70}{\us}. Furthermore, because we have to launch the kernel after the received data is ready, the GPU kernel launch overhead, which might be up to \qty{20}{\us}, is also added to end-to-end latency. All these additional latencies pose a significant overhead for \techname, which must rely on frequent network communications. 

To reduce such networking overhead, we develop a \textit{fully host-bypassed network} (FHBN) stack, which \textbf{completely eliminates host CPU involvement} in both control and data paths of GPU-aware networking. We describe how FHBN performs \texttt{send} and \texttt{recv} operations below.

\paragraph{FHBN recv.} To implement the FHBN recv function, we employ the \textit{device-side polling} technique to await the completion of the recv operation. Specifically, we allocate a \texttt{seqno} variable on the receiver's GPU memory. The sender increments the remote \texttt{seqno} with RDMA write after each send operation. The data send and \texttt{seqno} increment operations are batched in a single WR post and hence would not increase the end-to-end latency. When the receiver GPU is ready to receive and process the incoming data, it actively polls the value of \texttt{seqno} with a specialized GPU kernel. This approach not only eliminates the need for CPU involvement during the recv process, but also allows asynchronous launch of the polling kernel and subsequent computation kernels to the GPU stream. Therefore, the GPU kernel launch overhead is also removed from the critical path.

\begin{figure}[htbp]
    \centering
    \includegraphics[width=0.9\linewidth]{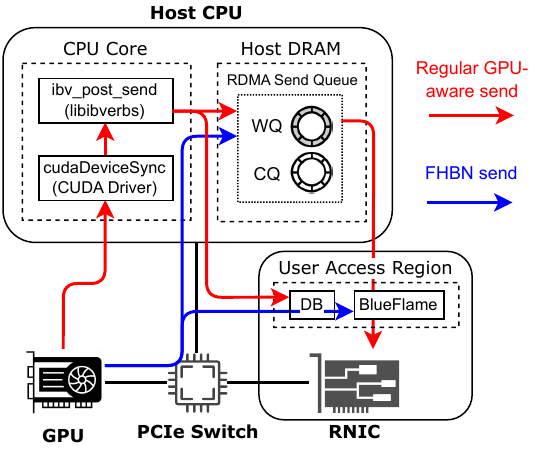}
    \caption{Diagram of WR submission with FHBN send and conventional GPU-aware send implementations.}
    \label{fig:fhbn-send}
\end{figure}

\paragraph{FHBN send.} The implementation of FHBN send, illustrated in \autoref{fig:fhbn-send}, is more involved as it necessitates the GPU to directly submit RDMA commands to RNIC. When the CPU submits a new RDMA WR to RNIC, it first enqueues the WR to the \textit{work queue} (WQ) in the host memory. Then, it tells the RNIC that there is outstanding work by ringing the \textit{doorbell} (DB), a special register in the \textit{user access region} (UAR) of the RNIC. The UAR is part of the RNIC's mmio region and is mapped to the address space of unprivileged applications to allow kernel-bypass RDMA operations. All above steps are implemented in the RDMA userspace library (\texttt{libibverbs}).

To enable direct RDMA command submission on GPUs, we have to allow GPUs to directly access the UAR via PCIe P2P. Specifically, we use the \texttt{cudaHostRegisterIoMemory} API to map the UAR into the GPU's address space. Then, we reimplement the RDMA command submission logic in CUDA device code. To further decrease latency, we leverage the BlueFlame mechanism, a hardware feature provided by Mellanox RNICs \cite{mellanox-prm}. This approach allows the WR to be directly submitted to the RNIC with mmio write to UAR, eliminating the need for the RNIC to fetch the WR from host memory via an expensive PCIe DMA read. Note that the WR should still be enqueued into the WQ, as the hardware may occasionally miss the BlueFlame WR and fall back to the regular workflow, particularly under heavy loads.

\subsection{Automated Model Converter}

\subsubsection{Model Splitting} \label{subsec:model-splitting}

In the attention offloading architecture, different operators of the LLM might be executed on different hardware; hence, we need to partition the model into slices, which is achieved by cutting at the attention operators. It often involves significant modifications to the existing codebase, primarily because the desired cutting points do not align with the LLM's inherent modular structure. This misalignment complicates the partitioning process and increases the risk of errors and inconsistencies within the heterogeneous system.

\begin{figure}[htbp]
    \centering
    \includegraphics[width=0.47\textwidth]{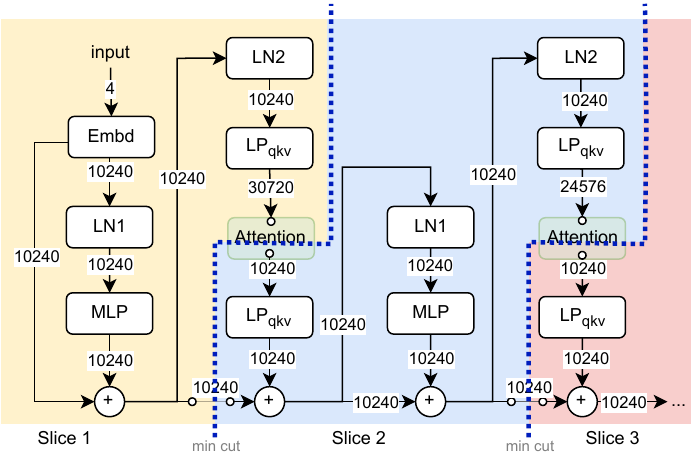}
    \caption{The partitioned computation graph of an LLM.}
    \label{fig:model-splitter}
\end{figure}

To facilitate model partitioning, we develop an automated model splitter capable of transforming the LLM into individually invokable slices, illustrated in \autoref{fig:model-splitter}. Given the LLM source code, the splitter uses symbolic execution to generate a weighted computation graph. The weight of each edge denotes the size of the data passed between the operators, which is derived from the model's shape specification. 

Due to the presence of residual connections and other intricate model constructs, directly removing the attention operator does not always result in a disconnected computation graph. Therefore, we compute the \emph{minimum weighted cut} of the remaining graph, from the input to the output of the attention operator. The edges in this minimum cut, representing the context that must be saved between slice invocations, are removed from the computation graphs. This process is iteratively applied to each attention operator, ultimately yielding $n+1$ model slices, where $n$ denotes the original number of the attention operators.

\subsubsection{Resource Utilization Overlapping} \label{subsubsec:converter:resource-overlapping}

While the attention operators and other operators in a transformer block are executed sequentially, a closer examination of the attention computation reveals the potential for achieving partial overlapping of resource utilization. Given an attention query $q$ and the set of token indices $I$, the attention computation can be carried out in a divide-and-conquer manner. Assume that $I$ can be written as the disjoint union of two subsets $I_1$ and $I_2$, and let 
\begin{align*}
A_q(I) &= \sum_{i \in I} \softmax \left(\frac{q^\top k_i}{\sqrt{d}}\right) v_i, \\
S_q(I) &= \sum_{i \in I} \exp \left( \frac{q^\top k_i}{\sqrt{d}} \right), 
\end{align*}
where $A_q(I)$ is the attention output and $S_q(I)$ is the denominator of $\softmax$, then $A_q(I)$ can be easily obtained by combining the partial attention results on $I_1$ and $I_2$, i.e., $[A_q(I_1), S_q(I_1)]$ and $[A_q(I_2), S_q(I_2)]$:
\[
A_q(I) = \frac{A_q(I_1) S_q(I_1) + A_q(I_2) S_q(I_2)}{S_q(I_1) + S_q(I_2)}.
\] 

\begin{figure}[htbp]
    \centering
    \begin{subfigure}{\linewidth}
        \includegraphics[width=\textwidth]{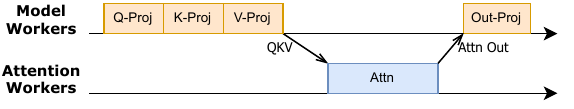}
        \caption{Without resource utilization overlapping.}
    \end{subfigure}
    \\ \bigskip
    \begin{subfigure}{0.92\linewidth}
        \includegraphics[width=\textwidth]{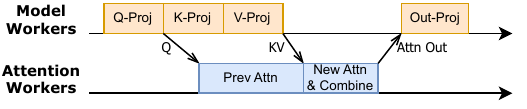}
        \caption{With resource utilization overlapping.}
    \end{subfigure} 
    \caption{Illustration of resource utilization overlapping by splitting the attention computation.}
    \label{fig:resource-utilization-overlapping}
\end{figure}

During LLM decoding, we may divide the current token set into two partitions during attention computation: all previous tokens (\text{prev}) and the newly generated token (\text{new}). Note that $[A_q(\text{prev}), S_q(\text{prev})]$ can be computed as soon as $q_n$ is ready; therefore, we may eagerly execute \textsf{Q-Proj} and transfer $q_n$, and then execute \textsf{K-Proj, V-Proj} and transfer $k_n, v_n$ to the attention workers. As illustrated in \autoref{fig:resource-utilization-overlapping}, this does not only improve the GPU utilization on both kinds of workers, but also reduces the end-to-end latency by hiding the communication behind the computation.

The above attention splitting optimization is integrated in our automated model converter. After dissecting the original model, the converter will generate a serial program of each model slice by computing a topological order of its computation graph. During this topological sort, we always put the \textsf{Q-Proj} operator and all its dependencies as early as possible. Then, we insert the ``send Q'' instruction immediately after the \textsf{Q-Proj} operator and ``send KV'' at the end of this slice. 

\subsection{Execution Pipelining}

Due to the serial nature of transformer-based models, if there is only one batch under processing, the memory device is idle when the computation device is working, and vice versa. To address this resource underutilization problem and increase system throughput, we may run multiple batches concurrently in a pipelined fashion. With properly designed pipelining, better hardware utilization can be achieved without sacrificing latency. We propose the rotational staggered pipelining to solve this problem.
%two pipelining schemes for this purpose. 

\begin{figure}[h]
    \centering
    %\begin{subfigure}{0.47\textwidth}
        \includegraphics[width=\linewidth]{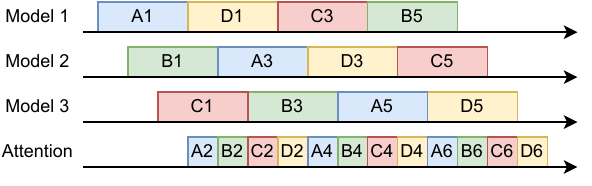}
        %\caption{Rotational staggered pipelining.} \label{fig:rotational-staggered-pipelining}
    %\end{subfigure}
    % \\ \bigskip
    % \begin{subfigure}{0.47\textwidth}
    %     \includegraphics[width=\textwidth]{figures/staggered-pipelining.pdf}
    %     \caption{Fixed staggered pipelining.} \label{fig:fixed-staggered-pipelining}
    % \end{subfigure} 
    \caption{Illustration of rotational staggered pipelining.}
    \label{fig:execution-pipelining}
\end{figure}

Assume that we execute $n$ batches concurrently. Let $t_m, t_a$ represent the time required for executing one model slice and one attention operator, respectively. As illustrated in \autoref{fig:execution-pipelining}, we deploy $n-1$ model replicas, with each replica starting its tasks at a time of $\frac{t_m}{n-1}$ later than the previous one. All batches share a common set of memory devices to maximize aggregated memory bandwidth and improve memory utilization. For every batch, the KV cache is evenly partitioned across these devices. All memory devices jointly compute the attention operator for a single batch. The number of memory devices is selected to make $t_a = \frac{t_m}{n-1}$. After the attention operator, each batch transitions to the next model replica according to a rotational schedule; that is, the $k$th model slice of the $j$th batch is executed on replica $(j+k) \bmod (n-1) + 1$. 

This rotational task scheduling, combined with the staggered execution intervals, guarantees seamless task transitions for each batch and ensures a conflict- and bubble-free workflow on each device. Furthermore, by increasing the number of concurrent batches, the overall inference latency can be reduced due to the decreased attention computation time. However, the rotational scheduling requires migrating batch execution contexts between computation devices. Note that when $n=2$, the context migration is unnecessary because both batches are executed within a single model replica.

\section{Implementation} \label{sec:implementation}
\systemname is implemented with \char`\~6000 lines of Python and C/C++ code, in addition to a few lines of CUDA code implementing custom kernels. The fully host-bypassed network stack is built on top of a modified version of rdma-core \cite{Linuxrdmardmacore}. \systemname uses Ray \cite{rayProductionizingScaling} to facilitate task scheduling and worker placement in distributed heterogeneous environments.

\paragraph{Fault tolerance.} With attention-offloading, we have two different types of accelerators. \systemname addresses faults in these two types of accelerators with different approaches. Note that all request states, i.e., the KV caches, are only stored in the attention devices. Consequently, should any model worker experience a failure, we can seamlessly replace that worker with a functioning one, without losing any progresses. In case of an attention worker failure, we reconstruct the KV cache by using the prompt texts and already generated tokens, which are stored in the LLM service front-end.

\paragraph{Handling the prefill-decode transition.}
During the prefill phase, the generated KV cache shall be transmitted to the attention workers for decoding. For each request, the global scheduler picks a set of model workers and attention workers to handle the decode phase. Like previous works \cite{patel2024splitwise,zhong2024distserve}, the KV cache is asynchronously transferred in a layer-by-layer fashion to hide the communication latency behind computation. Moreover, the data transfer is controlled by the attention workers: the attention workers only reads the KV cache from prefill workers during the free periods between receiving QKV tensors from model workers. This approach minimizes interference with ongoing decoding tasks.

\begin{figure}[htbp]
    \centering
    \begin{subfigure}[b]{3.504cm} % 4.38cm original 
        \includegraphics[width=\textwidth]{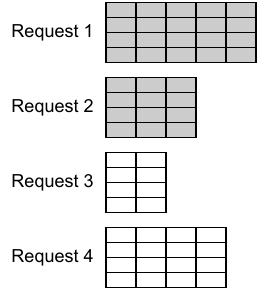}
        \caption{Request-level partition.}
    \end{subfigure}
    \hfill
    \begin{subfigure}[b]{4.720cm} % 5.90cm original
        \includegraphics[width=\textwidth]{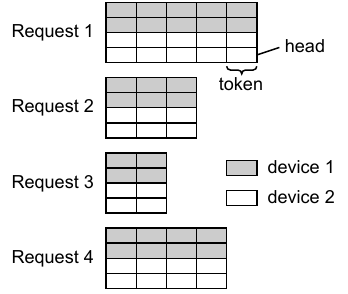}
        \caption{Head-level partition.}
    \end{subfigure}
    \caption{Work partition methods of the attention operator.}
    \label{fig:attention-work-partition}
\end{figure}
\paragraph{Attention parallelism.} Given the limited capability of a single device, we may use multiple memory devices to jointly store the KV caches and compute the attention operators. As depicted in \autoref{fig:attention-work-partition}, the attention operators can be parallelized among memory devices in various ways. One method is to distribute different requests across different devices; an alternative strategy is to partition and distribute the attention heads, which can also be computed independently, to different devices. The head-level partitioning approach ensures a balanced workload distribution, whereas the request-level partitioning may result in load imbalance due to the differences in sequence lengths and therefore the KV cache sizes among requests. However, head-level partitioning has limited flexibility, as it requires the number of memory devices to be divisible by the number of attention heads. We opt for head-level partitioning in \systemname, which offers optimal load balancing.

\section{Evaluation} \label{sec:evaluation} 
% \begin{figure*}[t]
%     \centering
%     \includegraphics[width=\textwidth]{figures/throughput.pdf}
%     \caption{Token generation throughput on LLaMA-13B and LLaMA-33B.}
%     \label{fig:throughput}
% \end{figure*}

\begin{figure*}[t]
    \centering
    \includegraphics[width=\textwidth]{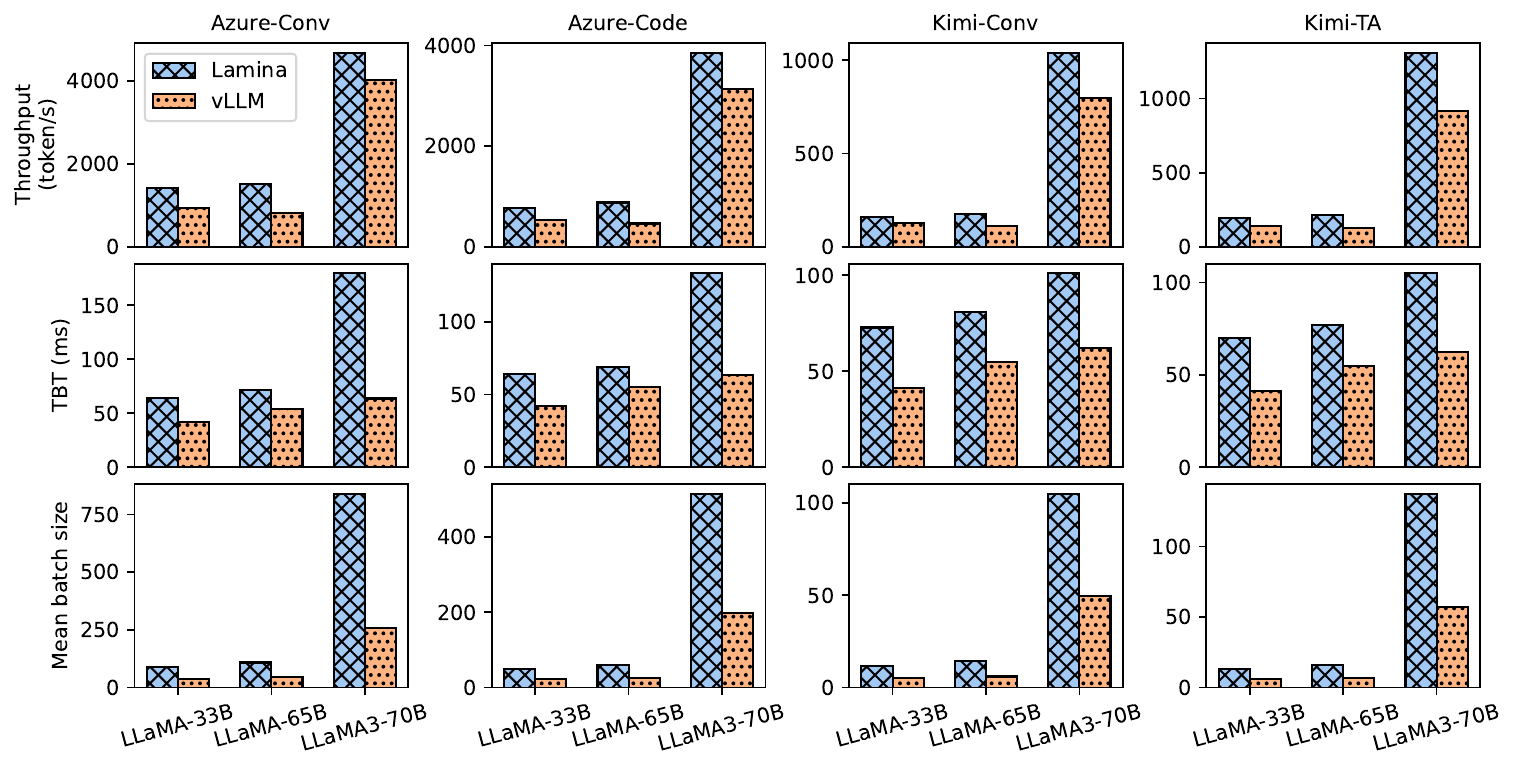}
    \caption{LLM decoding performance metrics of \systemname and vLLM, using hardware of approximately equal costs.}
    \label{fig:eval-equal-cost}
\end{figure*}

\begin{figure*}[t]
    \centering
    \includegraphics[width=\textwidth]{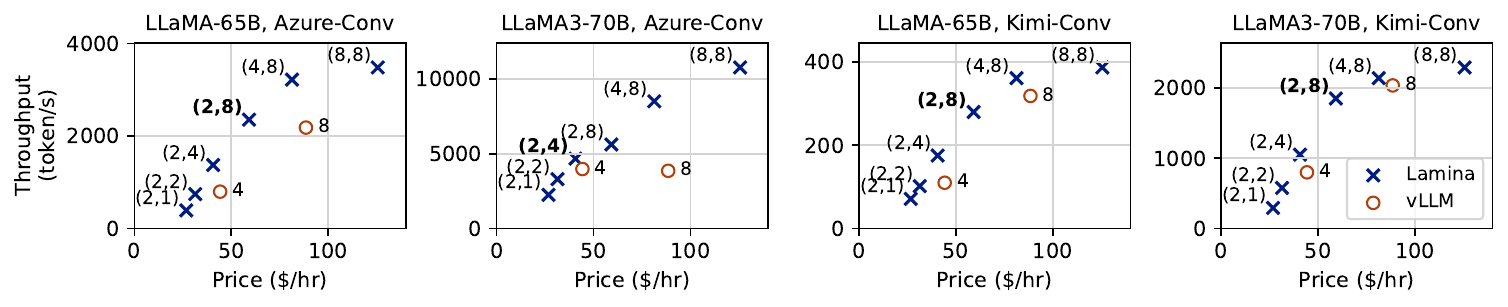}
    \caption{Decoding throughput and hardware cost with various hardware configurations. The DOPs for \systemname and tensor parallelisms for vLLM are annotated in the plot. The configuration with best cost efficiency is bolded.}
    \label{fig:eval-vary-config}
\end{figure*}

\paragraph{Testbed.} 
We deploy \systemname on a real heterogeneous cluster with two kinds of GPU nodes. Each node consists of either eight H100 or H20 GPUs, and each GPU is paired with a dedicated ConnectX-7 NIC via PCIe switch. The GPU nodes are interconnected with 400 Gbps RoCE network. We use H100 as compute-optimized GPUs and H20 as memory-optimized GPUs for \systemname.

\paragraph{Models.} \systemname supports a wide variety of LLM architectures, including OPT \cite{zhang2022opt}, LLaMA \cite{touvron2023llama}, and LLaMA3 \cite{llama3modelcard}. All these architectures have similar outlines and workload characteristics and only have minor differences irrelevant to system designs. Hence, as listed in \autoref{tab:model-summary}, we choose LLaMA-33B, LLaMA-65B, and LLaMA3-70B for evaluations. All model parameters and KV caches are stored in FP16 format.

\begin{table}[htbp]
\centering
\caption{Large language models used for evaluation.} \label{tab:model-summary}
\begin{tabular}{ccccc}
\toprule
\textbf{Model}   & \textbf{Parameters} & $L$ & $d$ & $G$ \\ \midrule
LLaMA-33B & 64.7 GB        & 60     & 6656  & 1      \\
LLaMA-65B & 130.1 GB        & 80     & 8192  & 1    \\
LLaMA3-70B & 137.5 GB        & 80     & 8192 & 8   \\ \bottomrule
\end{tabular}
\end{table}

\paragraph{Workloads} To mirror the real-world LLM use cases, we use four request traces collected from the production systems of two LLM service providers, Azure \cite{azuretrace2023,patel2024splitwise} and Kimi \cite{qin2024mooncake}. Due to data protection regulations, these traces only contain the sequence length of user requests but not the actual contents. Hence, we use requests of dummy tokens with the same sequence length for evaluation. The summaries of these traces, including the average prompt tokens ($l_{p}$) and average generated tokens ($l_{g}$), are listed in \autoref{tab:traces-summary}.

\begin{table}[htbp]
\centering
\caption{Request traces used for evaluation.} \label{tab:traces-summary}
\begin{tabular}{cccc}
\toprule
\textbf{Trace}  & \textbf{ \# Requests} & $l_{p}$ & $l_{g}$ \\ 
\midrule
Azure-Conv & 19366 & 1154.7 & 211.1  \\
Azure-Code & 8819 & 2047.8 & 27.9 \\
Kimi-Conv & 12031 & 12035.1 & 342.6 \\
Kimi-TA & 23608 & 8560.0 & 182.1 \\
\bottomrule
\end{tabular}
\end{table}

\paragraph{Baseline system.} We compare with vLLM \cite{vllm}, a state-of-the-art LLM serving system optimized for high throughput. vLLM also integrates optimizations from other LLM inference systems, such as continuous batching from Orca \cite{yu2022orca}. We use vLLM with homogeneous H100 GPUs and use tensor parallel for multi-GPU inference. As \systemname only focuses on the decode phase, we modify vLLM to remove the prefill phase during evaluation for a fair comparison.

\subsection{Serving Performance}

We evaluate the serving performance of \systemname against vLLM using real-world request traces. We first use homogeneous and heterogeneous hardware settings of similar costs, listed in \autoref{tab:equal-cost-hardware}, for vLLM and \systemname, respectively. Compared with vLLM, \systemname replaces half of the H100 devices to H20, which is cheaper but provides more memory capacity and bandwidth. We measure the token generation throughput, time between tokens (TBT), and average batch size. 

\begin{table}[h]
    \centering
    \caption{Equal-cost hardware configurations for evaluation.}
    \label{tab:equal-cost-hardware}
    \begin{tabular}{ccc} 
        \toprule
        \textbf{Model} & \textbf{\systemname} & \textbf{vLLM} \\ \midrule
        \multirow{2}{*}{LLaMA-33B} & DOP=(1,2) & $2\times$H100 \\
        & (\$20.32/hr) & (\$22.12/hr) \\
        \multirow{2}{*}{LLaMA-65B, LLaMA3-70B} & DOP=(2,4) & $4\times$H100 \\ 
        & (\$40.64/hr) & (\$44.24/hr) \\ \bottomrule
    \end{tabular}
\end{table}

As illustrated in \autoref{fig:eval-equal-cost}, \systemname consistently achieves $16.1 \sim 90.1\%$ higher throughput than vLLM among all models and traces, given comparable hardware costs. This enhancement is primarily attributed to the larger batch size attained by \systemname, which is $2.39\times$ of vLLM on average. These results demonstrate that \systemname effectively leverages the extra memory capacity provided by memory-optimized devices to boost decoding throughput. Note that the throughput and batch size of LLaMA3-70B is much larger than LLaMA-33B and LLaMA-65B; this is because LLaMA3-70B adopts GQA with a group size of 8, which results in a much smaller KV cache size per request.

\systemname experiences an increased token generation latency than vLLM. This can be attributed by two factors. First, \systemname adopts a larger batch size, which results in longer execution time on both model and attention workers. Second, the disaggregation of model and attention in \systemname may incur additional scheduling and networking overhead. Nevertheless, the end-to-end latency of \systemname can still meet the SLO requirements of interactive online LLM services in most cases.

We also explore the decoding throughput of \systemname and vLLM under various hardware configurations. Specifically, we adjust the DOPs for \systemname and the number of devices involving tensor parallelism for vLLM. As the results in \autoref{fig:eval-vary-config} shows, the throughput for \systemname rapidly increases with more attention workers added, which enables larger batch sizes. The addition of expensive model worker can only mildly improve the throughput by reducing the model-part execution latency. An exception is the LLaMA3-70B model, where the attainable batch size reaches 800 for $\DOP=(2,4)$, which already saturates the computation resources on model workers; hence, adding more memory devices will not dramatically improve the throughput. This indicates that the optimal ratio between model and attention workers varies for different models and workloads. In practice, we may conduct a performance profiling and select the best hardware configuration.

\begin{figure*}[t]
    \centering
    \includegraphics[width=\textwidth]{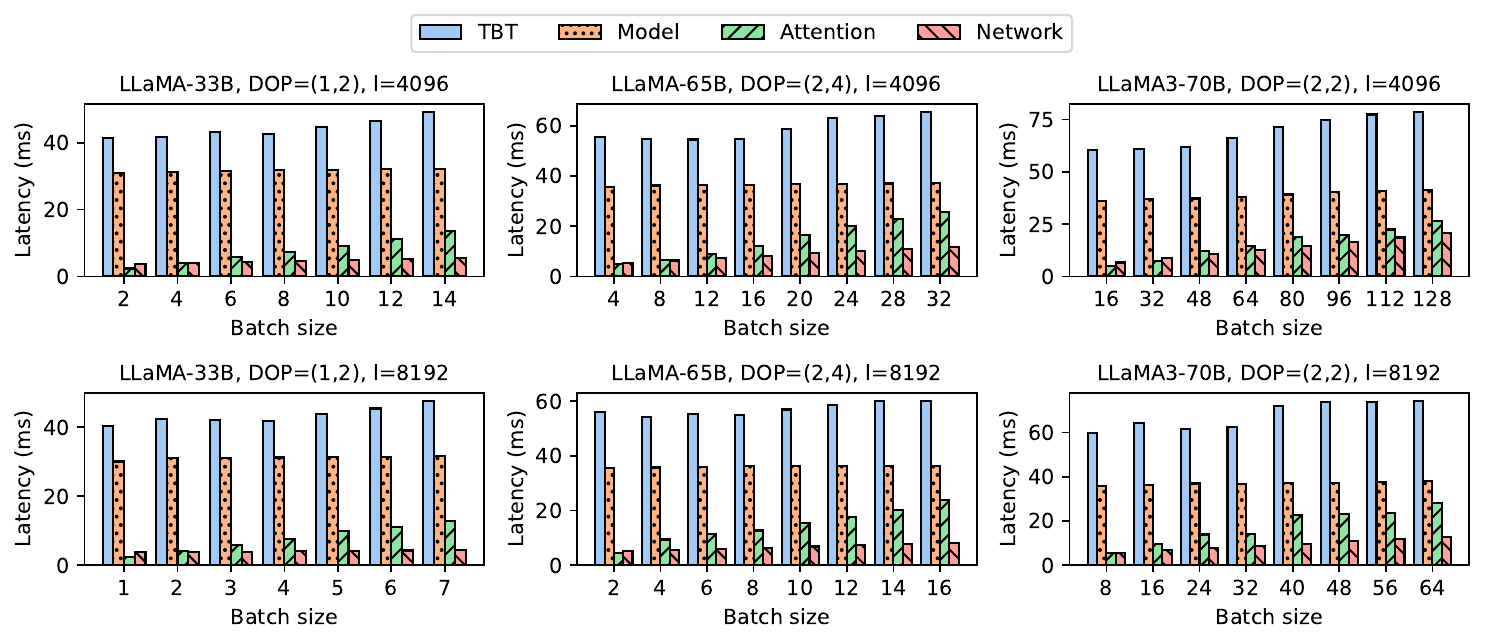}
    \caption{Token generation latency breakdown.}
    \label{fig:latency-breakdown}
\end{figure*}

\subsection{Latency Breakdown}

Latency is a crucial indicator of the service quality offered by LLM applications. In this subsection, we measure the time between tokens (TBT) across various system configurations, as well as the execution time for model and attention workers and the networking overhead. We use requests with fixed sequence lengths (4096 or 8192) as the workload and disable rotational pipelining to better reveal the time breakdown. 

As we can see from \autoref{fig:latency-breakdown}, for smaller batch sizes, the model execution time dominates the token generation latency. The attention and networking latency rapidly increases for larger batch sizes, while the model execution time remains almost constant. This indicates that the computation resource utilization gets improved as batch size increases. Note that the observed TBT might be less than the sum of model worker time, attention worker time, and network time. This is due to the automated resource utilization overlapping optimization, which will be further profiled in \autoref{subsec:eval:resource-utilization}. 

% It is observed that \systemname exhibits slightly higher latency compared to vLLM. This increase can be attributed to the additional data transfer and synchronization overhead between computation and memory devices. For larger batch sizes where vLLM runs out of memory, \systemname spends more time on networking. The attention computation time also increases with larger batch sizes as well as longer context lengths. However, the attention time can be decreased by adding more computation devices.

% Despite the increase in token generation latency, it is important to note that \systemname offers significant advantages in terms of efficient resource utilization and cost-effectiveness by enabling larger batch sizes, making it a viable and attractive alternative to traditional LLM inference systems.

\subsection{Network Stack Optimizations}

We evaluate the effectiveness of our fully host-bypassed network (FHBN) stack with a microbenchmark. Specifically, we conduct a ping-pong test between two GPUs located on distinct nodes, using NCCL, NCCL without GPUDirect RDMA, Gloo, and FHBN as the networking engine. The initiator GPU sends a varying amount of data to the remote GPU. Upon receiving the complete data, the remote GPU immediately sends it back to the initiator. We measure the round-trip time from the initiator GPU's perspective, which encompasses the time interval from the completion of the kernel that generates the data for transmission to the start of the kernel that consumes the received data.

\begin{figure}[htbp]
    \centering
    \includegraphics[width=\linewidth]{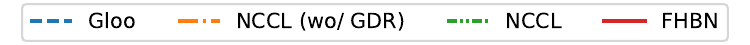}
    
    \begin{subfigure}[b]{0.48\linewidth}
        \includegraphics[width=\linewidth]{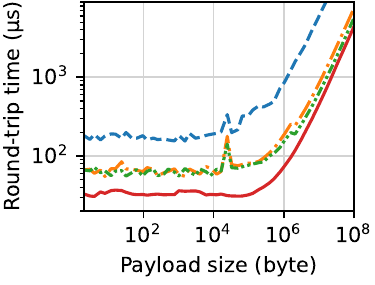}
        \caption{Round-trip time.}
    \end{subfigure}
    \hfill
    \begin{subfigure}[b]{0.48\linewidth}
        \includegraphics[width=\linewidth]{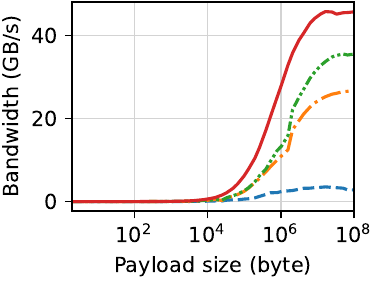}
        \caption{Bandwidth utilization.}
    \end{subfigure}
    \caption{Network ping-pong test between two GPUs on different nodes, interconnected with 400Gbps RoCE.}
    \label{fig:breakdown-network}
\end{figure}

As illustrated in \autoref{fig:breakdown-network}, for smaller data sizes, the round-trip time is primarily determined by network latency. In this case, FHBN achieves an end-to-end latency of \SI{33.0}{\us}, representing a 50.5\% reduction compared to NCCL's \SI{66.6}{\us} latency. This improvement is attributed to the removal of host CPU involvement in data transmission, eliminating expensive host-device synchronization and PCIe transactions. This improvement justifies the efficacy of our fully host-bypassed network stack design.

For larger payload sizes, the primary factor influencing networking time is the utilization of network bandwidth. In this scenario, FHBN reaches a peak network bandwidth of 45.7 GB/s, which corresponds to 91.4\% of the line rate. Conversely, NCCL only attains a bandwidth of 35.5 GB/s. As a result, FHBN can also serve as a superior alternative to existing communication libraries for point-to-point transmission of large GPU memory blocks within DCNs.

\subsection{Resource Utilization Overlapping} \label{subsec:eval:resource-utilization}

To assess the efficacy of resource utilization overlapping (\autoref{subsubsec:converter:resource-overlapping}) implemented in our automated model converter, we conducted a series of experiments on the LLaMA-65B and LLaMA3-70B models, with the optimization either enabled or disabled. We use request batches of varying sizes and the context length of each request is fixed at 4096.

% Llama-65B DOP=[2,2]
% Llama3-70B DOP=[2,2]
\begin{figure}[htbp]
    \centering
    \begin{subfigure}[b]{0.47\linewidth}
        \includegraphics[width=\linewidth]{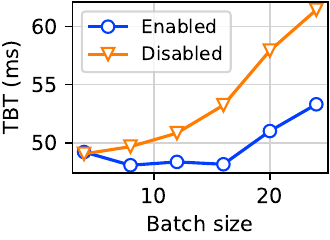}
        \caption{LLaMA-65B, DOP=$(2,2)$.}
    \end{subfigure}
    \hfill
    \begin{subfigure}[b]{0.50\linewidth}
        \includegraphics[width=\linewidth]{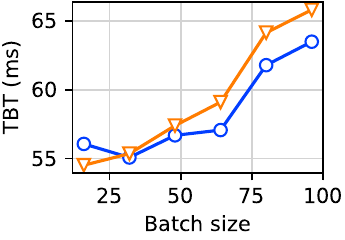}
        \caption{LLaMA3-70B, DOP=$(2,4)$.}
    \end{subfigure}
    \caption{Time between tokens (TBT) results with automatic resource utilization overlapping enabled and disabled.}
    \label{fig:results-overlapping}
\end{figure}

As illustrated in \autoref{fig:results-overlapping}, the LLaMA-65B model experiences a significant improvement in performance, achieving up to a 13.2\% with through automated resource utilization overlapping. The speedup is particularly notable for larger batch sizes, which produce larger KV tensors and result in greater latency reduction. The effectiveness is less pronounced for the LLaMA3-70B model, where the maximum latency reduction is only 3.5\%. This is because LLaMA3-70B adopts GQA, whose KV size is $8\times$ smaller. Consequently, there is less room for resource utilization overlapping in LLaMA3-70B.

\section{Discussion} \label{sec:discussion}
\paragraph{Generality of our techniques.} Although \systemname is built for \techname, relevant techniques can also be used to enable a wider range of fine-grained LLM disaggregation techniques in distributed heterogeneous environments. For example, LoRA \cite{hu2021loralowrankadaptationlarge} and Mixture-of-Experts (MoE) \cite{eliseev2023fastinferencemixtureofexpertslanguage,xue2024moeinfinityoffloadingefficientmoemodel} all add less computation-intensive operators to existing LLM architectures. Like \systemname, we may also offload the LoRA and MoE operators to less powerful but more economic remote accelerators to reduce the inference cost. Such operator-level disaggregations, unlike prefill-decode disaggregation, require frequent layer-wise communications and are considered not feasible unless an optimized networking stack like the one in \systemname is used. 
 
\paragraph{Alternative heterogeneous devices.} In \systemname, we may use more specialized accelerating devices for optimal performance and cost. For example, we anticipate that Processing-in-Memory (PIM) devices \cite{9749869,9251855,9895629,asifuzzaman2023survey,10.1145/3400302.3415640,9474146,10213232,10155455} will be a more suitable candidate for memory-optimized devices as they demonstrate even greater cost advantages alongside their larger capacity and higher bandwidth. Besides, we can also use CPU and DRAM for attention computation and KV cache storage. However, due to the relatively smaller bandwidth of host DRAM, it is preferable to also adopt sparse attention mechanisms \cite{xiao2024efficientstreaminglanguagemodels,beltagy2020longformer} to reduce the size of data read during attention computation.

\section{Related Work} \label{sec:related-work}

\paragraph{System optimizations for LLM Inference.} Splitwise \cite{patel2024splitwise} and DistServe \cite{zhong2024distserve} proposes prefill-decode disaggregation, which improves hardware utilization and minimizes the interference between the prefill and decode phases. Orca \cite{yu2022orca} proposes \textit{continuous batching}, that batches incoming requests in iteration granularity. Compared with whole-request batching, continuous batching greatly reduces resource waste caused by early termination during the decode phase. PagedAttention \cite{vllm} focuses on memory management optimizations, using fine-grained KV cache management to reduce memory waste. PagedAttention can also be used to optimize various decoding scenarios, like beam search and shared prefixes. These optimizations can all be used in our system. FlexGen \cite{flexgen} is a heterogeneous LLM inference system employing layer- and token-level task partitioning and scheduling. However, it does not account for the varying characteristics of different operators within a layer. LLM-tailored inference systems, like DeepSeed \cite{aminabadi2022deepspeed}, Megatron-LM \cite{shoeybi2020megatronlm}, and TensorRT-LLM \cite{tensorrt-llm}, use optimizations of various aspects including kernel optimization \cite{flashdecoding,hong2023flashdecodingpp}, advanced scheduling \cite{agrawal2023sarathi,turbotransformers,wu2023fast,li2023alpaserve}, and efficient memory management \cite{turbotransformers}.

\paragraph{Speculative Decoding} The speculative decoding technology \cite{leviathan2023fast, miao2023specinfer, liu2023online} enables parallel generation of multiple tokens for a single request during the decoding phase. This is done by \emph{guessing} the next few tokens using a smaller auxiliary model. These predicted tokens are then validated by the primary LLM. This validation of the predicted tokens can be executed in parallel, thereby enhancing the arithmetic intensity and reducing latency. However, speculative decoding can lead to a trade-off in throughput due to the auxiliary model's overhead and the potential need for re-execution in case of misprediction.

% \paragraph{} FrugalGPT \cite{chen2023frugalgpt} introduction a framework that uses strategies like prompt adaptation, LLM approximation, and LLM cascade to reduce the cost of using LLMs, focusing on adaptively triage diferent models.
% Tabi \cite{10.1145/3552326.3587438} uses a level of efficient DNN to initially process queries, returning results directly for those it handles with high confidence, achieving a balance between speed and accuracy without the need for extensive customization of the models. 

% \paragraph{} Flash-LLM\cite{xia2023flashllm} proposes leveraging sparsity through model pruning, which selectively removes less important parameters to reduce the model size without significantly impacting performance. 

\paragraph{Variations of the Attention Operator.} Researchers have developed many variations of the attention operator for large language models to mitigate the memory bottleneck. GQA \cite{ainslie2023gqa} and MLA \cite{deepseekai2024deepseekv2} are two recent attention mechanisms targeted for memory efficiency.  Model quantization uses reduced-precision formats (e.g., FP8) to store KV caches. Various sparse attention mechanisms  \cite{beltagy2020longformer,kitaev2020reformer,child2019generating,roy2021efficient,ye2019bp,qiu2019blockwise,liu2023ring,liu2023deja} have been adopted, focusing on a subset of all history key-value pairs during attention computation. All these modifications to the attention operator, however, might compromise the model quality.

\section{Conclusion} \label{sec:conclusion}
In this paper, we present \techname, an innovative architectural approach to improve the efficiency of LLM decoding. This approach is motivated by the observation that the LLM decoding phase can be divided into computation-intensive parts and memory-intensive parts (i.e., the attention operators). Hence, we may use computation- and memory-optimized devices for each part to improve the hardware resource utilization. Moreover, by adjusting the  To realize this idea, we design a revamped latency-optimized networking stack that facilitate the frequent data transfer between remote GPUs. We also develop automated tools for transforming and optimizing existing LLMs for \techname. We develop and deploy \systemname on a cluster comprising heterogeneous GPUs. Evaluation on traces collected from production systems show that \systemname provides $16.1 \sim 90.1\%$ higher throughput than heterogeneous solutions with similar hardware costs.

\bibliographystyle{plain}
\bibliography{references}

\end{document}